# AMI-Net+: A Novel Multi-Instance Neural Network for Medical Diagnosis from Incomplete and Imbalanced Data

Zeyuan Wang[1,2], Josiah Poon[1], and Simon Poon[1*]

[1] The University of Sydney, Sydney 2006, Australia
zwan7221@uni.sydney.edu.au,
{josiah.poon, simon.poon}@sydney.edu.au
[2] Medicinovo Inc., Beijing 100071, China
http://www.medicinovo.com/

**Abstract.** In medical real-world study (RWS), how to fully utilize the fragmentary and scarce information in model training to generate the solid diagnosis results is a challenging task. In this work, we introduce a novel multi-instance neural network, AMI-Net+, to train and predict from the incomplete and extremely imbalanced data. It is more effective than the state-of-art method, AMI-Net. First, we also implement embedding, multi-head attention and gated attention-based multi-instance pooling to capture the relations of symptoms themselves and with the given disease. Besides, we propose various improvements to AMI-Net, that the cross-entropy loss is replaced by focal loss and we propose a novel self-adaptive multi-instance pooling method on instance-level to obtain the bag representation. We validate the performance of AMI-Net+ on two real-world datasets, from two different medical domains. Results show that our approach outperforms other baseline models by a considerable margin.

## 1 Introduction

Worldwide, real-world study (RWS) has gained wide attention in recent years. However, when utilizing real-world data for studies, there are two main concerns [1]. First, data is always incomplete, since patients wouldn't perform all examinations in hospital, only necessary and required ones instead. Then, in most cases, the number of patients is far less than the number of healthy people, so from machine learning perspective, the dataset is always imbalanced.

For dealing with the incomplete data, imputation-based techniques are the most common, in which feature vectors are normally in a high-dimensional space and missing values are imputed by assumptions, such as zeros, mean of the feature, k-nearest-neighbors (k-NN) based and Expectation-Maximization (EM) based values [2-4]. However, redundant features and inaccurate assumptions for missing values always prevent the classifiers from achieving the better performance [5]. In this case, we address this problem in an alternative way that each patient's record is viewed as a sentence with a bag of words, i.e., symptoms. Through embedding, each word is represented as an instance, a dense vector. A computational model is developed after to select the most informative instances and generate the diagnosis results based on them. This strategy avoids to make assumptions for incomplete data and focuses on the instance-level information to screen out a number of invalid features. It is also known as multi-instance learning (MIL) [6].

Multi-instance learning is first proposed for drug molecule activity prediction [7], in which training data is organized as a set of labeled bags and each bag is represented by a list of unlabeled instances. Through learning, the MIL models allow to classify new bags in terms of their containing instances. In most previous MIL studies, instances are pre-given or from manually designed instance generator, such as EM-DD, mi-SVM, mi-Graph and miFV [8-11]. While, for achieving better flexibility and tractability, multi-instance neural networks (MINN) have been proposed in recent years [12-15] and applied in many domains such as image classification, video annotation and text categorization [16-18]. In MINN, the generation of instance representation and the following learning process will be accomplished by the model itself. Also, various computational modules, such as attention mechanism which has been widely applied in image and text analysis [19, 20], can endow MINNs with the ability to automatically uncover not only the relations between bags and their containing instances, but also the relations among instances themselves [21].

In our proposed method, AMI-Net+, we use AMI-Net [22] as our underlying architecture, which is also a MINN with the following effective computational layers, multi-head attention [23], gated attention-based multi-instance pooling [13] and a set of fully connected layers. Besides, we propose a novel self-adaptive multi-instance pooling method on instance level to obtain the bag representation. For dealing with the imbalanced data, we propose to utilize the focal loss [24] instead of the common used cross-entropy. Focal loss is proposed in the object detection community to solve the problem of extreme foreground-background class imbalance and it performs very well. During model training, it reduces the attention for well-classified candidates but to focus on the hard, less and misclassified ones. Inspired by this, focal loss is integrated in the neural network to make it much more robust to the imbalance data.

The next section in this paper gives a brief review of background and related works. Then we introduce the details of AMI-Net+, followed by experiments and conclusion.

## 2 Related Work

### 2.1 Multi-Instance Learning

Common supervised learning methods are to learn a mapping function $\Psi: X \rightarrow Y$ from the given dataset $\{(X_1, Y_1), (X_2, Y_2) \dots, (X_m, Y_m)\}$ where $X_i \in X$ is an instance, i.e., a feature vector and $Y_i \in Y$ is the corresponding label. Contrast to it, in MIL, the mapping function $\Psi$ is from $\{(X_1, Y_1), (X_2, Y_2) \dots, (X_m, Y_m)\}$ where $X_i \subseteq X$ is a bag of unlabeled instances $\{x_{i,1}, x_{i,2}, \dots, x_{i,n_i}\}$, $x_{i,j} \in \mathbb{R}^{d \times 1}$ and $Y_i \subseteq Y$ is a set of bag labels $\{y_{i,1}, y_{i,2}, \dots, y_{i,p_i}\}$, where $y_{i,q} \in \{0, 1\}$ [25].

For solving MIL problems, the standard assumption states that, for a specific class, the bag is labeled positive only if there contain one or more positive instances, otherwise, labeled negative. It can be formulated as follows:

$$y_{i,q} = \begin{cases} 1, & if\ \exists x_{i,j} \in X_i: \Psi(x_{i,j}) = 1 \\ 0, & otherwise \end{cases} \quad (1)$$

This is the underlying assumption for many early MIL methods [26, 9, 27], in which all witnesses are not necessary to be identified as long as a positive one found, the bag is labeled positive. While in this way, the correlation and distribution of instances are neglected. For solving this, a more general assumption is proposed, which is as follows. It is also under the basis property of MIL, permutation invariance [13].

$$\Psi(X_i) = \psi(\theta_{x_{i,j} \in X_{i,j}} \sigma(x_{i,j})) \quad (2)$$

where $\sigma$ is a suitable transformation and $\theta$ is a permutation invariance function, namely multi-instance pooling. About the $\psi$, it is a scoring function to obtain the bag score (or bag probability). According to the different choices of choosing $\psi, \theta$ and $\sigma$, MIL methods falls into two main categories [22]:

**Instance-Level Approach.** $\sigma$ is implemented on instance embeddings to compute the instance and bag representations. $\theta$ is the successive bag-level pooling for locating informative instances to calculate the bag score by a classifier $\psi$.

**Bag-Level Approach.** $\sigma$ is a classifier to compute the instance probabilities first. $\theta$ is then used for obtaining the bag score and $\psi$ is a simple linear transformation.

In this paper, we integrate these two approaches to fully utilize the instance-to-bag relationship. Moreover, due to flexibility left by the above MIL underlying function, we parameterize the contained transformations, i.e., MINN.

### 2.2 Multi-Instance Neural Network

The idea of using MINN is first proposed by Ramon and Raedt [28] to estimate the bag scores through the bag-level approach. MINNs take a various number of instances as input to learn instance and bag representations gradually. Thanks to the parameterization, the networks are optimized through the back-propagation [29]. In MINNs, a key component for bridging instances to bags is the multi-instance pooling layer, which is applied in instance-level or bag-level to obtain the bag representation or bag score.

There are mainly two ideas for choosing the multi-instance pooling methods, trainable ones or untrainable ones. Trainable ones are more efficient to discover hidden patterns such as attention-based multi-instance pooling [13] and dynamic pooling [15]. While, untrainable ones, including max pooling, mean pooling, sum pooling, etc., are more stable and flexible to implement. In our work, we adopt both methods on instance-level and bag-level respectively. In addition, when calculating the bag representation, the correlations of instances are essential to explore [30] and self-attention mechanism [23] shows its superiority in this respect.

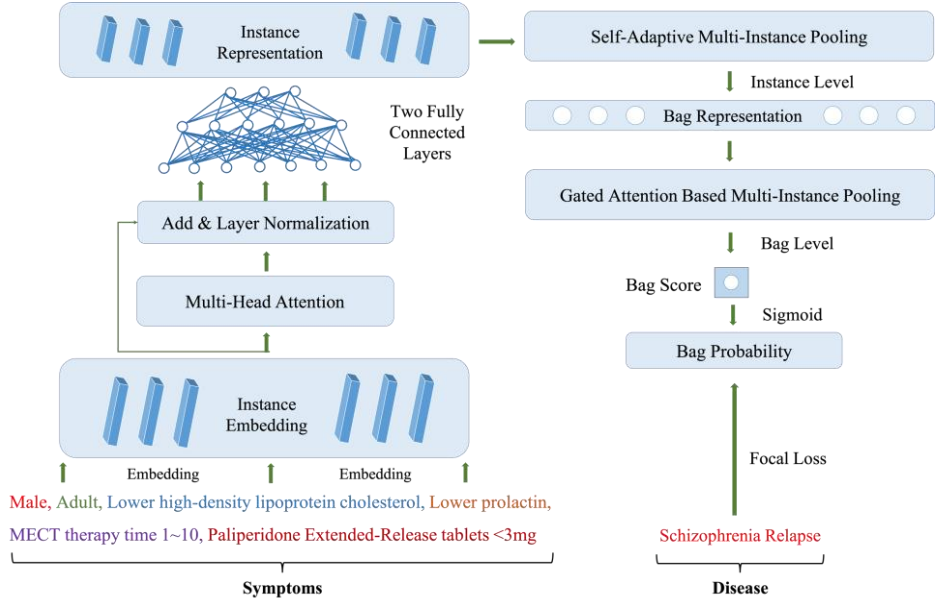

**Fig. 1.** The architecture of AMI-Net+

### 2.3 Self-Attention Mechanism

Self-attention is first proposed by Vaswani et. al [23] in the transformer architecture, to capture the correlations of words from the source and target sentences for the machine translation task. Their work demonstrates the validity of self-attention to reveal the syntactic and semantic information in text. In recent years, it has been applied in different real-life application such as semantic role labeling and biological relationship extraction [31, 32].

Motivated by this idea, we propose to consider symptoms (i.e., instances) as words and explore their relations in different subspaces, since symptoms are always correlated to each other in different parts of the body.

## 3 Methodology

### 3.1 Model Architecture

The overview of AMI-Net+ is shown in Fig. 1, that it takes a bag of symptoms (i.e., instances) as input and through embedding layer, each instance is mapped to a dense vector as the instance embeddings. The multi-head attention [23] is adopted after, followed by the layer normalization [33] and residual connection [34] to mine the instance correlations. Then we implement a set of fully connected layers to estimate the instance representations and the successive self-adaptive multi-instance pooling is on instance level to obtain the bag representation. For calculating the bag score, a gated attention-based multi-instance pooling is developed on bag level. At last, sigmoid and focal loss [24] is used for supervision.

### 3.2 Multi-Head Attention

Initially, multi-head attention takes three dense vectors as input, named *queries, keys and values*. It aims to map a *query* and a set of *key-value* pairs to the weighted sum of *values*, where the weights assigned to *values* are computed by the cosine similarity-based function with the *query* and corresponding *keys*. Moreover, in practical use, multi-head attention contains two main computational modules, scaled dot-product attention, multi-head transformation. The whole process is depicted as Fig. 2.

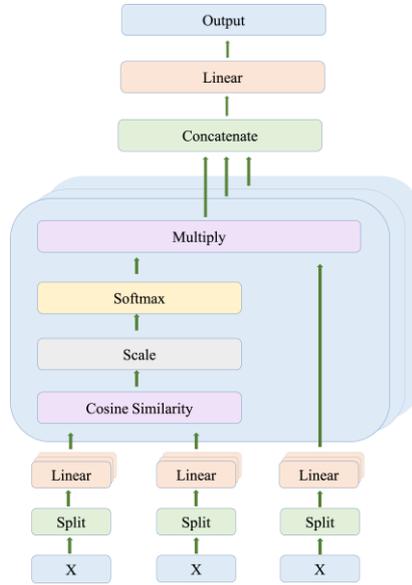

**Fig. 2.** The architecture of multi-head attention.

**Scaled Dot-Product Attention.** It takes the *query*, *keys* with $d_k$ dimensions, and values with $d_v$ dimensions as input and compute the cosine similarities, i.e., dot products, between the given *query* and all *keys* divided by a scaling factor $\sqrt{d_k}$. The scaling factor makes sure that the gradient in back propagation wouldn't vanish or be extreme small. Then a *softmax* function is applied after to compute the weights of *values*. Since in our method, we aim for extracting correlations of instances, so *queries*, *keys* and *values* are instances themselves and computation process is as follows:

$$sim(X, X) = \frac{X \cdot X^T}{\sqrt{d_k}} \quad (3)$$

$$Attention(X, X, X) = X \cdot softmax(sim(X, X)) \tag{4}$$

where $\cdot$ is element-wise product and $X$ is a bag of instance embeddings.

**Multi-Head Transformation.** Instead of performing attention in a single space, multi-head transformation splits the embedding dimensions into a number of representation subspaces to compute attention in different subspaces parallelly. Then these are concatenated together and after a linear projection, the final output is obtained, as shown in Fig. 3. It allows to jointly access information from different representation subspaces.

$$head_h = Attention(W^1 X, W^2 X, W^3 X) \tag{5}$$

$$MultiHead(X, X, X) = Concat_{h=1,2,\dots,H}\{head_h\}W^4 \tag{6}$$

Where $W^1, W^2, W^3 \in \mathbb{R}^{d_{model} \times d_k}$, $W^4 \in \mathbb{R}^{h d_k \times d_{model}}$, $H$ is the number of heads and $head_i$ represents $i^{th}$ subspace.

### 3.3 Self-Adaptive Multi-Instance Pooling

We propose a novel pooling method, self-adaptive multi-instance pooling, on instance-level to learn bag representation for successive classification. The input is instance representations and various untrainable pooling methods are applied to obtain a set of bag representations. Each one of them is considered as a view or a way to describe this bag. Inspired by the multi-view learning and ensemble learning, we concatenate all bag representations followed by a dense layer to calculate the weighted sum of different views. Let $X$ be a bag of $N$ instances with $K$ dimensions, and we formulate the proposed method as:

$$\forall_{j=1,2,\dots,N}: view_v = Pooling_{k=1,2,\dots,K}\{x_{j,k}\} \tag{7}$$

$$SelfAdaptive = Concat_{v=1,2,\dots,V}\{view_v\}W^v \tag{8}$$

where $W^v \in \mathbb{R}^{V \times 1}$, $x \in X$ and $V$ is the number of selected pooling methods. In this paper, we use max pooling, mean pooling, sum pooling and log-sum-exp pooling.

### 3.4 Gated Attention-Based Multi-Instance Pooling

In medical domain, the number of features is usually far more than the ones actually used. Therefore, how to allow the neural network to pay more attention on the instances most likely to be labeled as positive is an essential task. In our work, we propose to implement the gated attention-based multi-instance pooling on bag-level to locate the informative instances and explore the relations between instances and the given label. The process is as follows:

$$GateAttention = S \cdot Mean_{m=1,2,\dots,M}\{R_m\} \tag{9}$$

$$S = softmax(gate) \tag{10}$$

$$gate = W_1^T(tanh(RW_2) \cdot sigmoid(RW_3)) \tag{11}$$

where $W_2, W_3 \in \mathbb{R}^{M \times M}$, $W_1 \in \mathbb{R}^{M \times 1}$ and $R$ is the bag representation with $M$ embeddings from self-adaptive multi-instance pooling. The gate mechanism [35] is computed (See formula 11) for enhancing the expressiveness of the complex and non-linearity, which the $tanh$ function lacks.

In principle, gated attention-based pooling endows the model with the ability to assign different weights to instances within a bag, that also makes the model interpretable.

### 3.5 Focal Loss

For addressing the extreme imbalance problem, we propose to replace the cross-entropy loss with focal loss, which allows to guide the model focuses on the hard and misclassified samples [24]. After each feed forward, we can obtain the bag probability $y_{pred}$ and the corresponding bag label is $y_{true}$. In order to optimize our AMI-Net+, the focal loss is calculated as:

$$p_t = \begin{cases} y_{pred} & if\ y_{true} = 1 \\ 1 - y_{pred} & if\ y_{true} = 0 \end{cases} \quad (12)$$

$$FocalLoss = -\alpha(1 - p_t)^\gamma log(p_t) \quad (13)$$

where $\gamma \geq 0$ is the tunable focusing parameter, which reduces the loss contribution of easily classified samples, and $\alpha$ is a balance factor. In the experiments, we found that $\gamma = 2$ and $\alpha = 0.25$ achieved the best performance.

## 4 Experiments

### 4.1 Data Description

We evaluated the AMI-Net+ performance on two real-world medical datasets from traditional Chinese medicine (TCM) and western medicine (WM) respectively. In these two datasets, patients' symptoms are all standardized descriptive words or terms and they only have one disease. The examples are shown in Table 1.

**Traditional Chinese Medicine.** The TCM dataset is collected from clinical records of diabetic patients in a Chinese Medical Hospital in Beijing. There are 1617 patients and 186 different symptoms in the dataset. Each patient has a various number of symptoms, 1 at least and 17 at most. We aim for predicting whether they have meridian obstruction, a disease in TCM. Moreover, there are 1436 patients labeled negative and only 181 patients have this disease, so the dataset is extremely imbalanced.

**Western Medicine.** The WM dataset is collected by the Medicinovo Inc. in Beijing and all included are schizophrenia patients. The objective of our method is to predict whether their disease would recur within three months. In the dataset, 3927 patients are included and there are 88 medical features in total, such as married, high levels of prolactin and the total course is large than 3 years. For each patient, there are at most 21 features and 5 at least. Also, the dataset is extremely imbalanced that there exist only 224 positive labels out of 3927. The positive rate is only 0.057.

Table 1. Examples of TCM and WM datasets.

| Dataset | Features | Diagnosis |
|---|---|---|
| TCM | Urine color yellow, Sweat, Pruritus, Coldness of extremities, Perspiration | Meridian Obstruction |
| WM | Personal income 3000~5000, Unmarried, LOS<10 days, MECT<=1, Onset age<17, Total course<1095 days, Lorazepam tablets=0.5mg | Schizophrenia Relapse |

### 4.2 Experimental Setup

We first padded each record to the maximum length and embedded each medical feature or symptom to a dense vector with 512 dimensions. In multi-head attention, 4 and 8 heads were used on TCM and WM datasets respectively. Then the hidden sizes of two following fully connected layers were set to 256 and 128 respectively. About the final focal loss, we set the $\alpha$ and $\gamma$ to 0.25 and 2 and adam optimizer was applied to minimize it over the training data. In adam, we set the learning rate to 0.001, $\varepsilon$ to $1e^{-8}$ and the momentum parameters $\beta_1$, $\beta_2$ were set to 0.9 and 0.98 respectively. We used AUC, Accuracy, Precision and Recall as evaluation metrics. During the training process, the number of epochs was 200 and the batch size was 512. Moreover, the early stopping was utilized to select the best model in terms of the AUC score. For the fair comparison, we ran the experiments using 10-fold cross-validation with 5 repetitions.

In addition, about baseline models, we developed logistic regression (LR), support vector machine (SVM), random forest (RF), XGBoost (XGB), mi-Net, MI-Net, MI-Net with DS, MI-Net with RC, attention and gated attention based MINN (Att. Net, Gated Att. Net) [36-39, 14, 13]. Among them, LR, SVM, RF and XGB were all classic machine learning algorithms and constructed on the dataset in one-hot format with zero imputation. Also, the parameters of baseline models were tuned according to the AUC scores over the validation dataset.

## 5 Results and Analysis

### 5.1 Comparison with Baseline Models

The results of performance comparison with baseline models are shown in Table 2. Our proposed method performs best in terms of the AUC and recall scores, demonstrating its superiority in capture informative features of an extremely small number of positive samples. It is very vital in medical diagnosis, since for each patient, diseases cannot be missed. However, due to difficulties to collect sufficient positive samples, it is a challenging task. For instance, the two mainstream algorithms RF and XGB even cannot find any positive samples in the evaluation dataset. Moreover, according to Precision, AUC and Recall scores, MINNs (mi-Net, MI-Net, Att. Net etc.) perform much better than other classic machine learning methods (LR, SVM, RF and XGB), which demonstrates the better feasibility of MIL methods in many real-life applications, medical domain especially.

**Table 2.** Performance comparison on TCM and WM datasets.

| Models | TCM | | | | WM | | | |
|---|---|---|---|---|---|---|---|---|
| | AUC | Accuracy | Precision | Recall | AUC | Accuracy | Precision | Recall |
| LR | 0.760 | 0.944 | 0.200 | 0.017 | 0.755 | 0.882 | 0.396 | 0.116 |
| SVM | 0.657 | 0.946 | 0 | 0 | 0.703 | 0.889 | 0 | 0 |
| RF | 0.767 | 0.946 | 0 | 0 | 0.737 | 0.889 | 0 | 0 |
| XGBoost | 0.706 | 0.945 | 0.100 | 0.007 | 0.729 | 0.886 | 0.327 | 0.063 |
| mi-Net | 0.565 | 0.624 | 0.088 | 0.469 | 0.597 | 0.641 | 0.220 | 0.422 |
| MI-Net | 0.545 | 0.787 | 0.154 | 0.251 | 0.665 | 0.813 | 0.364 | 0.414 |
| MI-Net+DS | 0.510 | 0.621 | 0.045 | 0.383 | 0.586 | 0.731 | 0.358 | 0.290 |
| MI-Net+RC | 0.588 | 0.867 | 0.313 | 0.228 | 0596 | 0.861 | 0.353 | 0.358 |
| Att. Net | 0.608 | 0.849 | 0.342 | 0.143 | 0.642 | 0.861 | 0.368 | 0.244 |
| Gated Att. Net | 0.576 | 0.832 | 0.248 | 0.140 | 0.607 | 0.755 | 0.319 | 0.354 |
| AMI-Net | 0.702 | 0.907 | 0.356 | 0.283 | 0.702 | 0.818 | 0.399 | 0.468 |
| AMI-Net+ | 0.774 | 0.779 | 0.301 | 0.689 | 0.761 | 0.802 | 0.165 | 0.644 |

Worst Performance    Best Performance

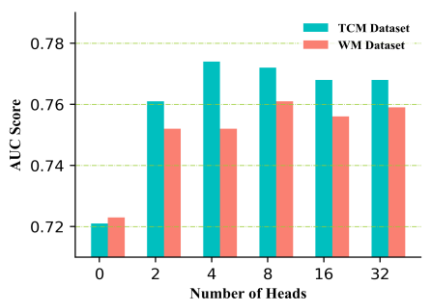

**Fig. 3.** Comparison of different number of heads in multi-head attention.

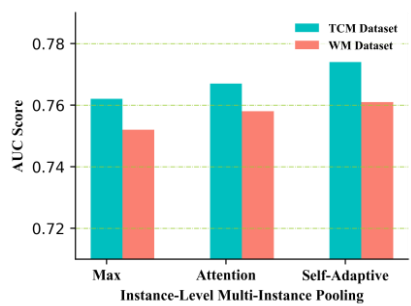

**Fig. 4.** Comparison of different multi-instance pooling methods on instance level

### 5.2 Comparison of Different Number of Heads

In this section, we aim for evaluating how different number of heads in the multi-head attention influence the model performance. The experiment was conducted on both datasets with 0, 4, 8, 16 and 32 heads, where 0 denoted that we didn't implement multi-head attention in our method. The model performance was measured by the AUC score. As shown in Fig. 3, the model without multi-head attention performs much worse than others, indicating the its necessity to capture the correlations of clinical features before classification.

In addition, the model with 8 heads is the best choice on WM dataset, meaning that clinical features in WM dataset mostly correlated to each other in 8 aspects. And on TCM dataset, when using 4 heads, the model explores symptom correlations most efficiently and performs the best. This experimental result is also consistent to the TCM knowledge, that TCM symptoms are all collected from four following methods, inspection, listening and smelling, inquiry and pulse-taking, representing four aspects of the body condition.

### 5.3 Comparison of Different Multi-Instance Pooling Methods

When using different multi-instance pooling methods on instance level, we tested how the model performed in terms of the AUC score. Since max pooling and attention-based pooling had been shown to be superior [14, 13], we used them as baselines. The results (See Fig. 4) show that our proposed pooling method demonstrates its efficacy and performs best. In addition, the max pooling behaves worst, indicating that capturing the information in one embedding dimension would be insufficient for representing an instance.

### 5.4 Evaluation of Focal Loss

For analyzing the behavior of focal loss (FL), we compared the model performance with it and cross-entropy loss (CE) on both datasets. As shown in Table 3, the model with FL is much better on identifying positive samples than CE. Although the Accuracy and Precision scores are lower than CE, in extremely imbalanced data, Recall score is more important, since if the full prediction is 0, the Accuracy score is still 0.946. In general, FL enables our approach simple and highly effective to solve the problem of extremely imbalanced data.

**Table 3.** Performance comparison of focal loss and cross-entropy loss.

| Loss | TCM | | | | WM | | | |
|---|---|---|---|---|---|---|---|---|
| | AUC | Accuracy | Precision | Recall | AUC | Accuracy | Precision | Recall |
| FL | 0.774 | 0.779 | 0.301 | 0.689 | 0.761 | 0.802 | 0.165 | 0.644 |
| CE | 0.746 | 0.863 | 0.391 | 0.394 | 0.707 | 0.939 | 0.398 | 0.204 |

🟥 Worst Performance  🟩 Best Performance

## 6 Conclusion

This paper attempts to solve the problem of incomplete and extremely imbalanced data with a novel multi-instance neural network, AMI-Net+, in which a multi-head attention and gated attention based multi-instance pooling method are applied to capture the correlations of symptoms and the informative ones. Also, a novel instance-level multi-instance pooling method is proposed to obtain the better bag representation. At last, the common used cross-entropy loss is replaced by the focal loss. The experimental results indicate that the proposed method performs much better than all other baseline models in terms of the AUC and Recall scores.

The study demonstrates the superiority and feasibility of AMI-Net+ in real-life medical applications and real-world studies.